\documentclass{article}

\usepackage{arxiv}

\usepackage[utf8]{inputenc} 
\usepackage[T1]{fontenc}    
\usepackage{hyperref} 
\usepackage{url}            
\usepackage{booktabs}       
\usepackage{amsfonts}       
\usepackage{amsmath}
\usepackage{nicefrac}       
\usepackage{microtype}      
\usepackage{lipsum}		
\usepackage{graphicx}
\usepackage{cite}
\usepackage{doi}
\usepackage{algorithm}
\usepackage{algpseudocode}
\usepackage{wrapfig}
\usepackage{multirow}
\usepackage{caption}
\usepackage{footnote}

\title{Self-Reflection in LLM Agents: Effects on Problem-Solving Performance}

\date{} 					

\author{
    Matthew Renze \\
    Johns Hopkins University\\
    \texttt{mrenze1@jhu.edu} \\
    \And
    Erhan Guven \\
    Johns Hopkins University\\
    \texttt{eguven2@jhu.edu} \\
}

\renewcommand{\headeright}{}
\renewcommand{\undertitle}{}
\renewcommand{\shorttitle}{Self-Reflection in LLM Agents}

\hypersetup{
    colorlinks=true,
    linkcolor=black,
    citecolor=black,
    urlcolor=blue,
    pdftitle={Self-Reflection in LLM Agents: Effects on Problem-Solving Performance},
    pdfauthor={Matthew Renze},
    pdfkeywords={large language model, LLM, agent, self-reflection},
}

\begin{document}
\maketitle

\begin{abstract}
	In this study, we investigated the effects of self-reflection in large language models (LLMs) on problem-solving performance. We instructed nine popular LLMs to answer a series of multiple-choice questions to provide a performance baseline. For each incorrectly answered question, we instructed eight types of self-reflecting LLM agents to reflect on their mistakes and provide themselves with guidance to improve problem-solving. Then, using this guidance, each self-reflecting agent attempted to re-answer the same questions. Our results indicate that LLM agents are able to significantly improve their problem-solving performance through self-reflection ($p < 0.001$). In addition, we compared the various types of self-reflection to determine their individual contribution to performance. All code and data are available on GitHub at \href{https://github.com/matthewrenze/self-reflection}{https://github.com/matthewrenze/self-reflection}
\end{abstract}


\section{Introduction}

\subsection{Background}

Self-reflection is a process in which a person thinks about their thoughts, feelings, and behaviors. In the context of problem-solving, self-reflection allows us to inspect the thought process leading to our solution. This type of self-reflection aims to avoid making similar errors when confronted with similar problems in the future. 

Like humans, large language model (LLM) agents can be instructed to produce a chain of thought (CoT) before answering a question. CoT prompting has been shown to significantly improve LLM performance on a variety of problem-solving tasks \cite{Kojima2022, Wei2022, Zhou2023}. However, LLMs still often make errors in their CoT due to logic errors, mathematical errors, hallucination, etc. \cite{Ji2022, Usman2023, Payandeh2023, Huang2023, Ji2023, Minaee2024}.

Also similar to humans, LLM agents can be instructed to reflect on their own CoT. This allows them to identify errors, explain the cause of these errors, and generate advice to avoid making similar types of errors in the future \cite{Shinn2023, Pan2023, Madaan2023, Toy2024, WangY2023, Asai2023}. 

Our research investigates the use of self-reflection in LLM agents to improve their problem-solving capabilities.

\subsection{Prior Literature}

Over the past few years, we've seen the emergence of AI agents based on LLM architectures \cite{Wang2023, Xi2023}. These agents have demonstrated impressive capabilities in solving multi-step problems \cite{Yao2022, WangG2023, Shinn2023}. In addition, they've been observed successfully using tools, including web browsers, search engines, code interpreters, etc. \cite{Nakano2021, WangG2023, Shinn2023, Schick2023}.

However, these LLM agents have several limitations. They have limited knowledge, make errors in reasoning, hallucinate output, and get stuck in unproductive loops \cite{Ji2022, Usman2023, Payandeh2023, Huang2023, Ji2023, Minaee2024}.

To improve their performance, we can provide them with a series of cognitive capabilities. For example, we can provide them with a CoT \cite{Kojima2022, Wei2022, Zhou2023}, access to external memory \cite{Lewis2020, Gao2023, Zhong2023B, WangZ2024}, and the ability to learn from feedback \cite{Yao2022, Shinn2023, WangG2023}.

Learning from feedback can be decomposed into several components. These components include the source of the feedback, the type of feedback, and the strategy used to learn from feedback \cite{Pan2023}. There are two sources of feedback (i.e., internal or external feedback) and two main types of feedback (i.e., scalar values or natural language) \cite{Pan2023, Madaan2023}. 

There are also several strategies for learning from feedback. These strategies depend on where they occur in the LLM's output-generation process. They can occur at model-training time, output-generation time, or after the output has been generated. Within each of these three phases, there are various techniques available (e.g., model fine-tuning, output re-ranking, and self-correction) \cite{Pan2023}.

In terms of learning from self-correction, various methods are currently being investigated. These include iterative refinement, multi-model debate, and self-reflection \cite{Pan2023}.

Self-reflection in LLM agents is a metacognitive strategy also known as introspection \cite{Toy2024, WangY2023}. Some research studies have indicated that LLMs using self-reflection are able to identify and correct their mistakes \cite{Madaan2023, Shinn2023, Ji2023, Asai2023}. Others have indicated that LLMs cannot identify errors in their reasoning; regardless, they still may be able to correct them with external feedback \cite{Huang2023, Tyen2023}.

\subsection{Contribution}

Our research builds upon the prior literature by determining which aspects of self-reflection are most beneficial in improving an LLM agent's performance on problem-solving tasks. It decomposes the process of self-reflection into several components and identifies how each component contributes to the agent's overall increase in performance.

In addition, it provides insight into which types of LLMs and problem domains benefit most from each type of self-reflection. These include LLMs such as GPT-4, Llama 2 70B, and Gemini 1.5 Pro. It also includes various problem domains such as math, science, medicine, etc.

This information is useful to AI engineers attempting to build LLM agents with self-reflection capabilities. In addition, it is valuable to AI researchers studying metacognition in LLM agents.

\section{Methods}

\subsection{Data}
Our test dataset consists of a set of multiple-choice question-and-answer (MCQA) problems derived from popular LLM benchmarks. These benchmarks include ARC, AGIEval, HellaSwag, MedMCQA, etc. \cite{Clark2018, Zellers2019, Pal2022, Zhong2023A, Liu2020, Wang2021}.

We preprocessed and converted these datasets into a standardized format. Then, we randomly selected 100 questions from each of the ten datasets to create a multi-domain exam consisting of 1,000 problems.

For a complete list of the source problem sets used to create the MCQA exam, see Table \ref{tab:exams}. For a sample of an MCQA problem, see Figure \ref{fig:mcqa-problem} in the appendix.

\begin{table*}[ht]
    \caption{Problem sets used to create the 1,000-question multi-domain MCQA exam.}
    \label{tab:exams}
    \vskip 0.15in
    \begin{center}
        \begin{small}
            \begin{tabular}{llllrl}
                \toprule
                Problem Set & Benchmark & Domain & Questions & License & Source \\
                \midrule
                ARC Challenge Test & ARC & Science & 1,173 & CC BY-SA & \cite{Clark2018} \\
                AQUA-RAT & AGI Eval & Math & 254 & Apache v2.0 & \cite{Zhong2023A} \\
                Hellaswag Val & Hellaswag & Common Sense Reasoning & 10,042 & MIT & \cite{Zellers2019} \\
                LogiQA (English) & AGI Eval & Logic & 651 & GitHub & \cite{Zhong2023A, Liu2020} \\
                LSAT-AR & AGI Eval & Law (Analytic Reasoning) & 230 & MIT & \cite{Zhong2023A, Wang2021} \\
                LSAT-LR & AGI Eval & Law (Logical Reasoning) & 510 & MIT & \cite{Zhong2023A, Wang2021} \\
                LSAT-RC & AGI Eval & Law (Reading Comprehension) & 260 & MIT & \cite{Zhong2023A, Wang2021} \\
                MedMCQA Valid & MedMCQA & Medicine & 6,150 & MIT & \cite{Pal2022} \\
                SAT-English & AGI Eval & English & 206 & MIT & \cite{Zhong2023A} \\
                SAT-Math & AGI Eval & Math & 220 & MIT & \cite{Zhong2023A} \\
                \bottomrule
            \end{tabular}
        \end{small}
    \end{center}
    \vskip -0.1in
\end{table*}
{\footnotesize Note: The GitHub repository for LogiQA does not include a license file. However, both the paper and readme.md file states that "The dataset is freely available."}

\subsection{Models}

We evaluated our agents using nine popular LLMs, including GPT-4, Llama 2 70B, Gemini 1.5 Pro, etc. \cite{Anthropic2024A, Anthropic2024B, Cohere2024A, Cohere2024B, Pichai2023, GeminiTeam2023, Pichai2024, GeminiTeam2024, OpenAI2022, OpenAIGPT35, OpenAI2023A, OpenAI2023B, Meta2023, Touvron2023, MistralAI2024}. All models were accessed via cloud-based APIs hosted by Microsoft, Anthropic, and Google.

Each of these LLMs has its own unique strengths and weaknesses. For example, LLMs like GPT-4, Gemini 1.5 Pro, and Claude Opus are powerful LLMs with a large number of parameters \cite{OpenAI2023B, GeminiTeam2024, Anthropic2024B}. However, they have a significantly higher cost per token than smaller models like GPT-3.5 and Llama 2 7B \cite{OpenAIGPT35, Touvron2023}.

For a complete list of LLMs used in our experiment, see Table \ref{tab:models}.

\begin{table*}[ht]
    \caption{LLMs used in the experiment.}
    \label{tab:models}
    \vskip 0.15in
    \begin{center}
        \begin{small}
            \begin{tabular}{lllrl}
                \toprule
                Name & Vendor & Released & License & Source \\
                \midrule
                Claude 3 Opus & Anthropic & 2024-03-04 & Closed & \cite{Anthropic2024A, Anthropic2024B} \\
                Command R+ & Cohere & 2024-04-04 & Open & \cite{Cohere2024A, Cohere2024B} \\
                Gemini 1.0 Pro & Google & 2023-12-06 & Closed & \cite{Pichai2023, GeminiTeam2023} \\
                Gemini 1.5 Pro (Preview) & Google & 2024-02-15 & Closed & \cite{Pichai2024, GeminiTeam2024} \\
                GPT-3.5 Turbo & OpenAI & 2022-11-30 & Closed & \cite{OpenAI2022, OpenAIGPT35} \\
                GPT-4 & OpenAI & 2023-03-14 & Closed & \cite{OpenAI2023A, OpenAI2023B} \\
                Llama 2 7B Chat & Meta & 2023-07-18 & Open & \cite{Meta2023, Touvron2023} \\
                Llama 2 70B Chat & Meta & 2023-07-18 & Open & \cite{Meta2023, Touvron2023} \\
                Mistral Large & Mistral AI & 2024-02-26 & Open & \cite{MistralAI2024} \\                
                \bottomrule
            \end{tabular}
        \end{small}
    \end{center}
    \vskip -0.1in
\end{table*}

\subsection{Agents}

We investigated eight types of self-reflecting LLM agents. These agents reflect upon their own CoT and then generate self-reflections to use when attempting to re-answer questions. Each of these agents uses a unique type of self-reflection to assist it. We also included a single non-reflecting (i.e., \textit{Baseline}) agent as our control.

Listed below are the various types of agents and the type of self-reflection they generate and use to re-answer questions:

\begin{itemize}
    \item \textbf{Baseline} - no self-reflection capabilities.
    \item \textbf{Retry} - informed that it answered incorrectly and simply tries again.
    \item \textbf{Keywords} - a list of keywords for each type of error.
    \item \textbf{Advice} - a list of general advice for improvement.
    \item \textbf{Explanation} - an explanation of why it made an error.
    \item \textbf{Instructions} - an ordered list of instructions for how to solve the problem.
    \item \textbf{Solution} - a step-by-step solution to the problem.
    \item \textbf{Composite} - all six types of self-reflections.
    \item \textbf{Unredacted} - all six types without the answers redacted
\end{itemize}

The Baseline agent is our control for the experiment and a lower bound for the scores. It informs us how well the base model answers the question without using any self-reflection. The Baseline agent used standard prompt-engineering techniques, including domain expertise, CoT, conciseness, and few-shot prompting \cite{Bsharat2023, Renze2024B, Kojima2022, Wei2022, Zhou2023}. The sampling temperature was set to 0.0 for all LLMs to improve reproducibility \cite{Renze2024A}. See Figure \ref{fig:answer-prompt} in the appendix for an example of the Baseline answer prompt.

The self-reflecting agents used the same prompt-engineering techniques as the Baseline agent to re-answer questions. However, they also reflected upon their mistakes before attempting to re-answer. While re-answering, the self-reflection was injected into the re-answer prompt to allow the agent to learn from its mistakes. See Figures \ref{fig:reflection-prompt} and \ref{fig:re-answer-prompt} in the appendix for examples of the self-reflection prompt and the re-answer prompt.

We redacted all of the answer labels (e.g., "A", "B", "C") and answer descriptions (e.g., "Baltimore", "Des Moines", "Las Vegas") from the agents' self-reflections. However, the \textit{Unredacted} agent retains this information. This agent is only used to provide an upper bound for the scores. Essentially, the Unredacted agent tells us how accurately the LLM could answer the questions when given the correct answer in its self-reflection.

\subsection{Process}

\begin{figure}[h]
    \begin{minipage}{0.53\textwidth}   
        \centering
        \includegraphics[width=\linewidth]{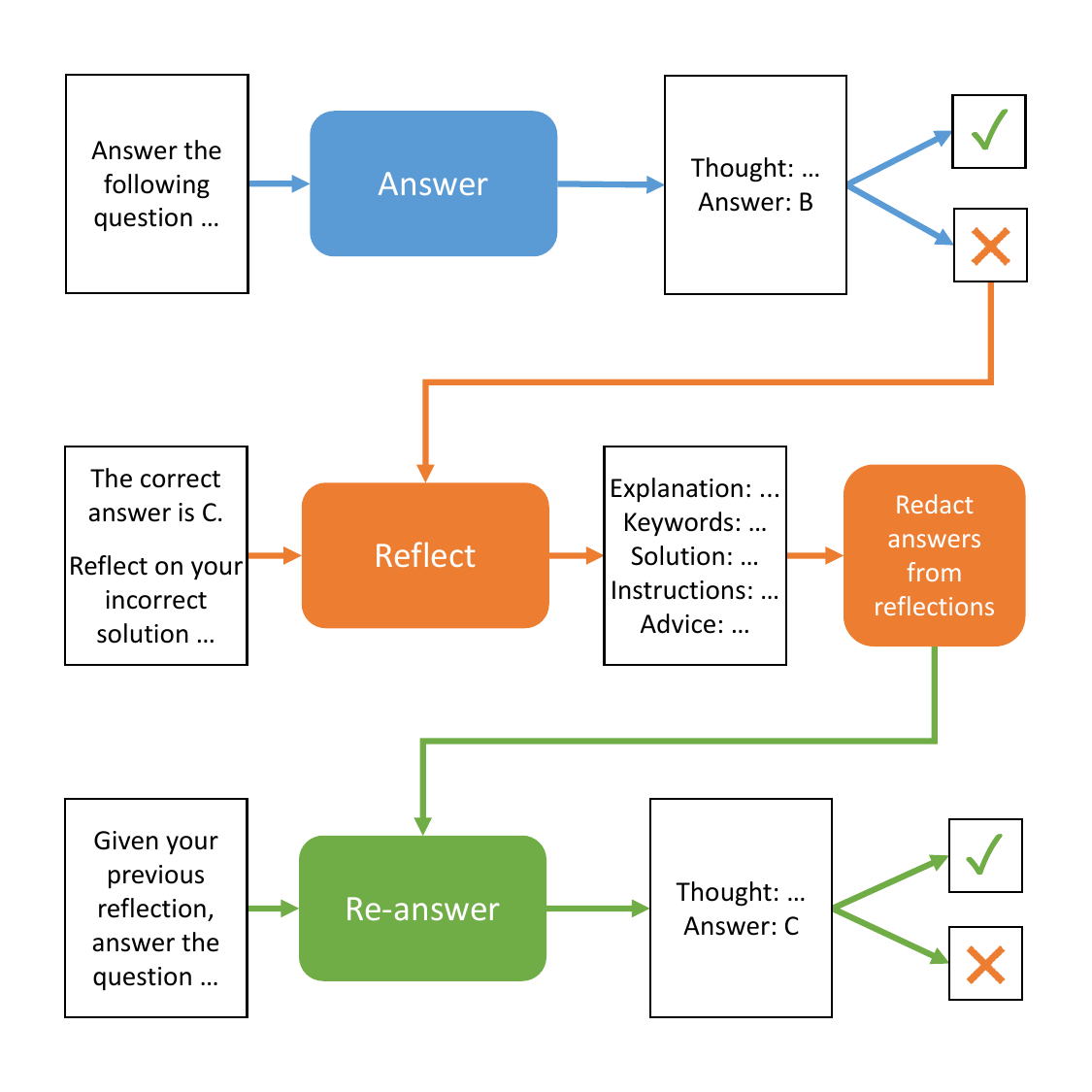}
        \caption{Diagram of the self-reflection experiment.}
        \label{fig:self-reflection-diagram}
    \end{minipage}
    \hfill
    \begin{minipage}{0.45\textwidth}
        \begin{algorithm}[H]
            \caption{Self-reflection Experiment (Batch)}
            \label{alg:self-reflection-algorithm}
            \begin{algorithmic}[1]
                \For{each model, exam, and problem}
                    \State Create the answer prompt 
                    \State Answer the question
                    \If{the answer is incorrect}
                        \State Add the problem to the incorrect list
                    \EndIf
                \EndFor
                \State Calculate the Baseline agent scores
                \Statex
                \For{each model, exam, and problem}
                    \State Reflect upon the incorrect solution
                    \State Generate the self-reflections
                    \If{not the Unredacted agent}
                        \State Redact the answers 
                    \EndIf
                    \State Separate the reflections by type
                \EndFor
                \Statex
                \For{each model, agent, exam, and problem}
                    \State Create the re-answer prompt
                    \State Inject the agent's reflection
                    \State Re-answer the question
                \EndFor
                \State Calculate the reflected agent scores
            \end{algorithmic}
        \end{algorithm}
    \end{minipage}
\end{figure}

First, the Baseline agent answered all 1,000 questions. If a question was answered correctly, it was added to the Baseline agent's score. If it was answered incorrectly, it was added to a queue of incorrectly answered questions to be reflected upon (see Figure \ref{fig:self-reflection-diagram}).

Next, for each incorrectly answered question, the self-reflecting agents reflected upon the problem, their incorrect solution, and the correct answer. Using the correct answer as an external feedback signal, they each generated one of the eight types of self-reflection feedback described above.

Then, a find-and-replace operation was performed on the text of each self-reflection to redact the answer labels and answer descriptions. For example, we replaced answer labels (e.g., "A", "B", "C") and answer descriptions (e.g., "Baltimore", "Des Moines", "Las Vegas") with the text "[REDACTED]".\footnote{The process we used to redact answer labels and descriptions was greedy. It often redacted additional text that did not leak the answer. However, we felt it necessary to err on the side of caution by eliminating any possible answer leakage.} This was done to all of the self-reflecting agents, except for the \textit{Unredacted} agent, to prevent answer leakage in the self-reflections.\footnote{It is important to note that the self-reflections generated by the Explanation, Instructions, and Solution agents indirectly leak information about the correct answer without directly specifying the correct or incorrect answers. However, they generated this information on their own based on nothing more than being provided the correct answer during the self-reflection process.}

Finally, for each incorrectly answered question, the self-reflecting agents used their specific self-reflection text to assist them in re-answering the question. We calculated the scores for all agents and compared them to the Baseline agent for analysis.

While LLM agents typically operate over a series of iterative steps, the code for this experiment was implemented as batch operations to save time and cost. So, each step in the self-reflection process occurred in one of four batch phases described above. Conceptually, the experiment represented \textit{virtual} multi-step agents. However, the technical implementation of the experiment was actually a series of batch operations (see Algorithm \ref{alg:self-reflection-algorithm}).

\subsection{Metrics}
We used correct-answer accuracy as our primary metric to measure the performance of the agents. Accuracy is calculated by dividing the number of correctly answered questions by the total number of questions.

However, to reduce the cost of running our experiment, we did not have the self-reflecting agents re-answer all of the questions that were correctly answered by the Baseline agent. Rather, the self-reflecting agents only re-answered the incorrectly answered questions. We then added the self-reflecting agent's correct re-answer score to the Baseline agent's score to create a new total score for the self-reflecting agent. 

The calculations for accuracy used in our experiment are listed in Equation(s) \ref{eq:accuracy}. In these equations, the subscript $_{base}$ refers to the Baseline agent's correct-answer score, and the subscript $_{ref}$ is the reflection agent's correct re-answer score.

\begin{align}    
    \text{Accuracy}_{\text{base}} &= \frac{\text{Correct}_{\text{base}}}{\text{Total}_{\text{base}}}
      & \hspace{2cm} 
    \text{Accuracy}_{\text{ref}} &= \frac{\text{Correct}_{\text{base}} + \text{Correct}_{\text{ref}}}{\text{Total}_{\text{base}}} 
    \label{eq:accuracy}
\end{align}

\subsection{Analysis}
When comparing the scores of the self-reflecting agents to the Baseline agent, we performed the McNemar test to determine statistical significance and report p-values. This test was specifically chosen because our analysis compared two series of binary outcomes (i.e., correct or incorrect answers). These outcomes were paired question-by-question across both the Baseline agent and self-reflecting agent being compared.

The McNemar test compares the number of discordant pairs in the two sets of pair-wise outcomes. To compute the test statistic, we create a $2 \times 2$ contingency table of the outcomes. In cell $a$, we state the number of cases where both agents answered incorrectly. Cell $d$ contains the cases where they both answered correctly. Cell $b$ contains incorrect-correct answer pairs and cell $c$ contains correct-incorrect answer pairs (which, in our case, will always be zero) \cite{McNemar1947}.

The McNemar's test statistic is calculated as:
\[
\chi^2 = \frac{(b-c)^2}{b+c} \quad \text{where } b \text{ and } c \text { are the discordant pairs in } 
\left[ \begin{array}{cc}
a & b \\
c & d \\
\end{array} \right]
\]

\section{Results}

\subsection{Performance by Agent}

\begin{figure}[h]
    \centering
    \includegraphics[width=\linewidth]{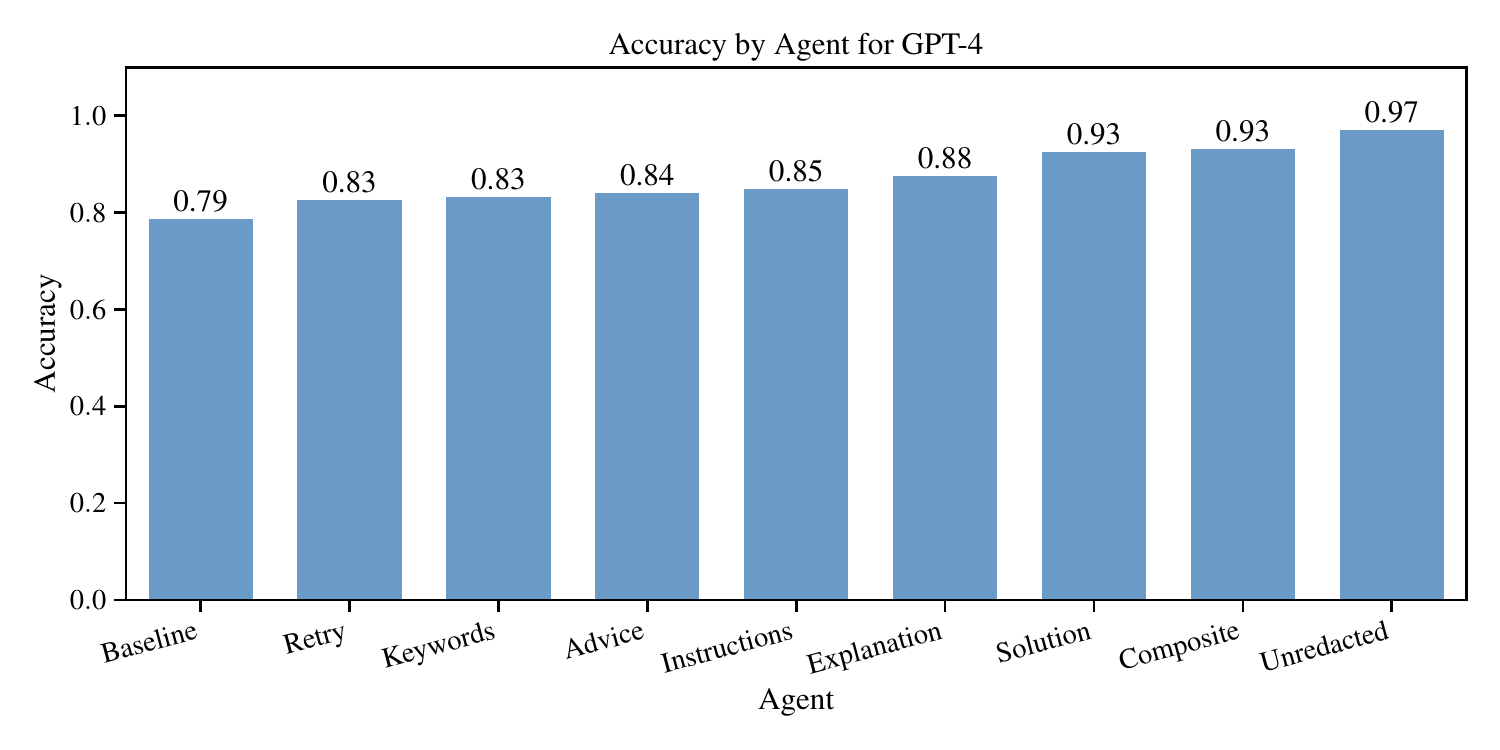}
    \caption{All self-reflection types improved the accuracy of GPT-4 agents.}
    \label{fig:accuracy-by-agent}
\end{figure}

Our analysis revealed that agents using various types of self-reflection outperformed our Baseline agent. The increase in performance was statistically significant ($p < 0.001$) for all types of self-reflection across all LLMs. We can use GPT-4 as an example case. In Figure \ref{fig:accuracy-by-agent}, we can see that all types of self-reflection improve the accuracy of the agent in solving MCQA problems. See Table \ref{tab:accuracy-by-agent-for-gpt-4} in the appendix for a numerical analysis of the results for GPT-4.

\subsection{Performance by Model}

\begin{figure}[h]
    \centering
    \includegraphics[width=\linewidth]{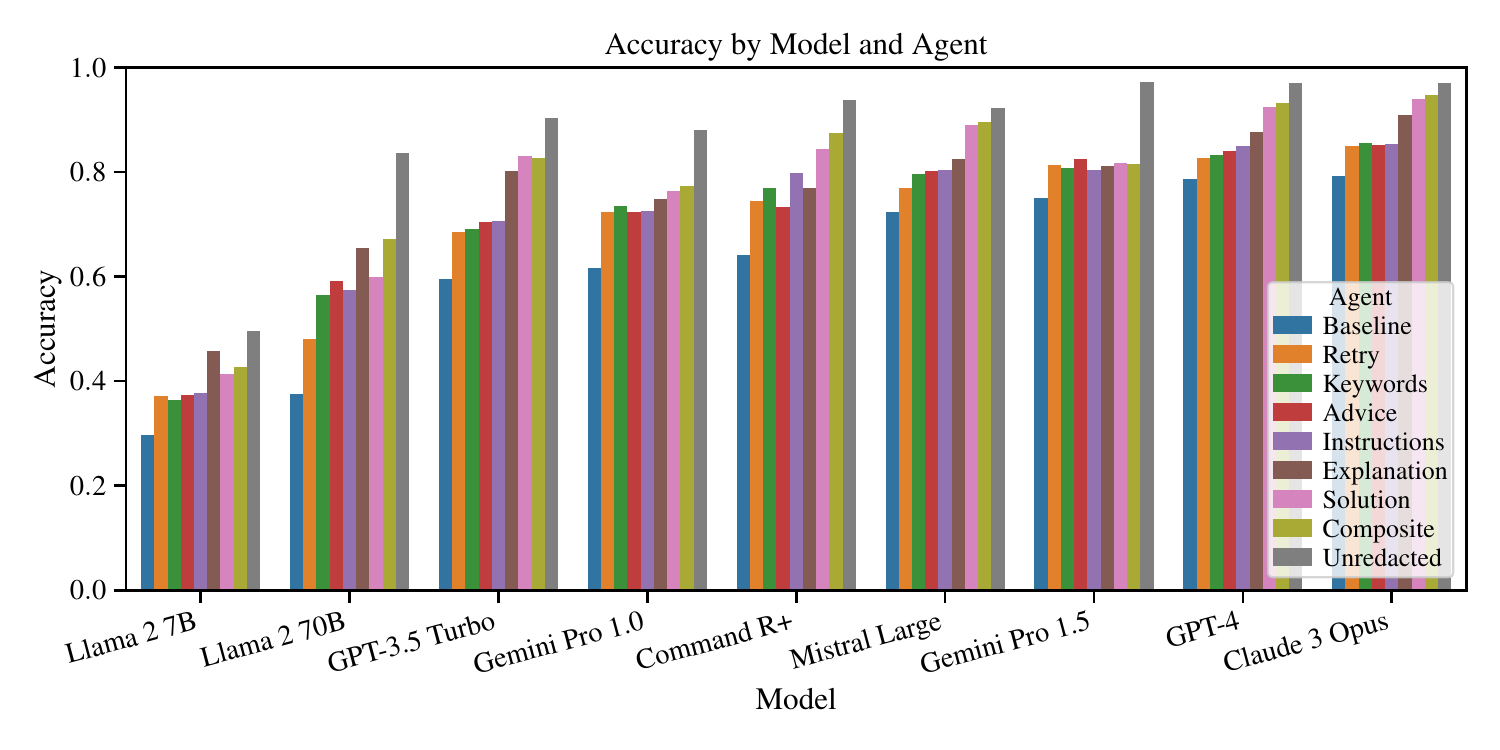}
    \caption{All LLMs we tested showed a similar pattern of improvement across self-reflection agents.}
    \label{fig:accuracy-by-model-and-agent}
\end{figure}

In terms of performance by model, every LLM that we tested demonstrated similar increases in accuracy across all types of self-reflection. In all cases, the improvement in performance was statistically significant ($p < 0.001$). See Figure \ref{fig:accuracy-by-model-and-agent} for a plot of accuracy by model and agent. See Table \ref{tab:accuracy-by-model-and-agent} for a numerical analysis of accuracy across all models.

\subsection{Performance by Exam}
\begin{figure}[h]
    \centering
    \includegraphics[width=\linewidth]{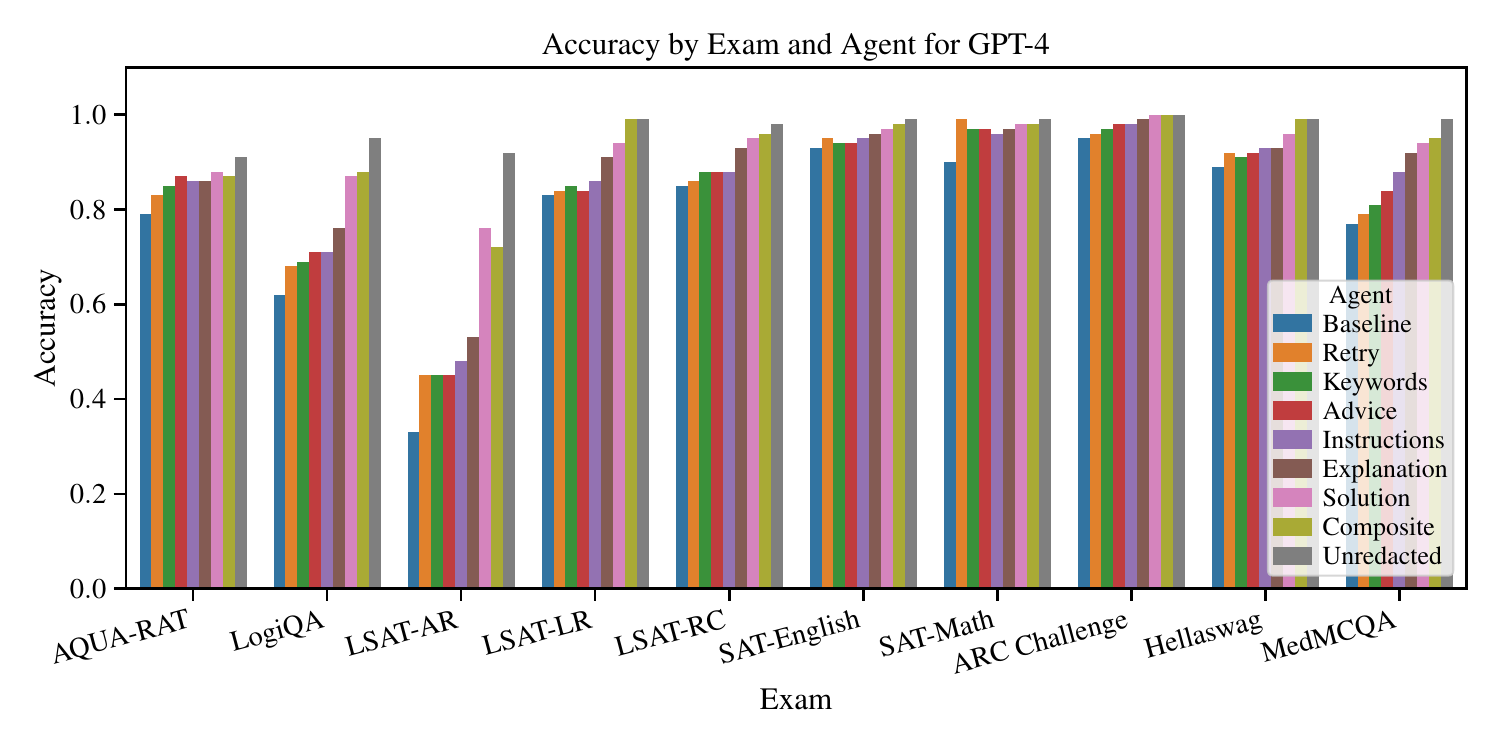}
    \caption{The increase in performance from self-reflection was larger for some exams and smaller for others.}
    \label{fig:accuracy-by-exam-and-agent}
\end{figure}

In terms of performance by exam, we saw that self-reflection significantly increased performance for some problem domains. However, other problem domains were less affected. For example, we saw the largest improvement on the LSAT-AR (Analytical Reasoning) exam. Other exams, like the SAT English exam, had much smaller effects. See Figure \ref{fig:accuracy-by-exam-and-agent} for a plot of accuracy by exam and agent for GPT-4. See Table \ref{tab:accuracy-by-agent-and-exam-for-gpt-4} in the appendix for a numerical analysis of the results.

\section{Discussion}

\subsection{Interpretation}

Based on these results, all types of self-reflection improve the performance of LLM agents. In addition, these effects were observed across every LLM we tested. Self-reflections that contain more information (e.g., Instructions, Explanation, and Solution) outperform types of self-reflection with limited information (e.g., Retry, Keywords, and Advice).

The difference in accuracy between the self-reflecting agents and the Unredacted agent demonstrates that we were effectively eliminating direct answer leakage from the self-reflections. However, the structure of feedback generated by the Instruction, Explanation, Solution, and Composite agents clearly provides indirect guidance toward the correct answer without directly giving the answer away.

Interestingly, the Retry agent significantly improved performance across all LLMs. As a result, it appears that even the mere knowledge that the agent previously made a mistake improves the agent's performance while re-answering the question. We hypothesize that this is either the result of the agent being more diligent in its second attempt or choosing the second most likely answer based on its re-answer CoT. Further investigation will be required to answer this question.

\subsection{Limitations}

First, the LLM agent we created for this experiment only solved a single-step problem. The real value in LLM agents is their ability to solve complex multi-step problems by iteratively choosing actions that lead them toward their goal. As a result, this experiment does not fully demonstrate the potential of self-reflecting LLM agents.

Second, API response errors may have introduced a small amount of error into our results. API errors typically occurred when content-safety filters were triggered by the questions being asked. In most cases, this may have amounted to an error in reporting an agent's accuracy of less than 1\%. However, in the case of the Gemini 1.0 Pro and Mistral Large models, this error could be as high as 2.8\%.

Third, the top-performing LLMs scored above 90\% accuracy for most exams. As a result, the increase in scores for the top exams was compressed near the upper limit of 100\% (i.e., a perfect score). This compression effect makes it difficult to accurately assess the performance increase. As a result, our analysis would benefit from exams with a higher level of difficulty. 

Finally, for all models and agents, the LSAT-AR (Analytical Reasoning) exam was the most difficult and also the most benefited by self-reflection. This large increase in performance from a single exam had the potential to skew the aggregate results across all exams. Using a set of exams with more uniform difficulty would eliminate this skewness.

\subsection{Implications}

Our research builds upon prior work on LLM agents and self-reflection.

It has practical implications for AI engineers who are building agentic LLM systems. Agents that can self-reflect on their own mistakes based on error signals from the environment can learn to avoid similar mistakes in the future. This will also help prevent the common issue of agents getting stuck in unproductive loops because they continue repeating the same mistake indefinitely.

In addition, our research has theoretical implications for AI researchers studying metacognition in LLMs. If LLMs are able to self-reflect on their own CoT, other similar metacognitive processes may also be leveraged to improve their performance.

\subsection{Future Research}

First, we recommend repeating this experiment using a more complex set of problems. Using problems as difficult or more difficult than the LSAT-AR exam would better reflect the performance improvement from self-reflection by avoiding compression of the scores around 100\% accuracy.

Second, we recommend performing an experiment using multi-step problems. This would allow the agents to receive feedback from their environment after each step to use as external signals for error correction. It would also demonstrate the potential of self-reflection on long-horizon problems.

Third, we recommend repeating this experiment while providing the agents with access to external tools. This would allow us to see how error signals from the tools benefit self-reflection. For example, we could observe how an agent adapts to compiler errors from a Python interpreter or low-rank search results from a search engine.

Fourth, we recommend repeating this experiment with agents that possess external memory. Having an agent answer the same questions based on self-reflection is only beneficial from an experimental standpoint. Real-world agents need to store self-reflections and retrieve them (using Retrieval Augmented Generation) when encountering similar but not necessarily identical problems.

Finally, we recommend a survey of self-reflection across a wider set of LLMs, agent types, and problem domains. This would help us better characterize the effects of self-reflection and provide further empirical evidence for the potential benefits of self-reflecting LLM agents.

\section{Conclusion}

In this study, we investigated the effects of self-reflection in LLM agents on problem-solving tasks. Our results indicate that LLMs are able to reflect upon their own CoT and produce guidance that can significantly improve problem-solving performance. These performance improvements were observed across multiple LLMs, self-reflection types, and problem domains. This research has practical implications for AI engineers building agentic AI systems as well as theoretical implications for AI researchers studying metacognition in LLMs.

\section{Acknowledgements}
Funding for this research was provided by \href{https://www.microsoft.com/}{Microsoft} and the \href{https://renzeai.org/}{Renze AI Research Institute}.

\bibliographystyle{IEEEtran}
\bibliography{references}

\appendix

\newpage
\section{Appendix}
\subsection{Results}

\begin{table}[h!]
\caption{Comparison of accuracy by agent for GPT-4}
\label{tab:accuracy-by-agent-for-gpt-4}
\centering
\begin{tabular}{lrrrrr}
\hline
\textbf{Agent Name} & \textbf{Accuracy} & \textbf{Difference} & \textbf{Test Statistic} & \textbf{p-value} \\ 
\hline
Baseline & 0.786 & N/A & N/A & N/A \\
Retry & 0.827 & 0.041 & 39.024 & $< 0.001$ \\
Keywords & 0.832 & 0.046 & 44.022 & $< 0.001$ \\
Advice & 0.840 & 0.054 & 52.019 & $< 0.001$ \\
Instructions & 0.849 & 0.063 & 61.016 & $< 0.001$ \\
Explanation & 0.876 & 0.090 & 88.011 & $< 0.001$ \\
Solution & 0.925 & 0.139 & 137.007 & $< 0.001$ \\
Composite & 0.932 & 0.146 & 144.007 & $< 0.001$ \\
Unredacted & 0.971 & 0.185 & 183.005 & $< 0.001$ \\ 
\hline
\end{tabular}
\end{table}

\begin{table}[h!]
\centering
\caption{Accuracy by model and agent}
\label{tab:accuracy-by-model-and-agent}
\fontsize{7pt}{9pt}
\begin{tabular}{lrrrrrrrrr}
\hline
\textbf{Model Name} & \textbf{Baseline} & \textbf{Retry} & \textbf{Keywords} & \textbf{Advice} & \textbf{Instruction} & \textbf{Explanation} & \textbf{Solution} & \textbf{Composite} & \textbf{Unredacted} \\ 
\hline
Claude 3 Opus & 0.792 & 0.849 & 0.855 & 0.852 & 0.853 & 0.908 & 0.939 & 0.947 & 0.971 \\
Cohere Command R+ & 0.641 & 0.745 & 0.770 & 0.733 & 0.798 & 0.770 & 0.843 & 0.874 & 0.937 \\
Gemini 1.0 Pro & 0.617 & 0.724 & 0.734 & 0.724 & 0.725 & 0.748 & 0.763 & 0.774 & 0.881 \\
Gemini 1.5 Pro & 0.751 & 0.813 & 0.807 & 0.824 & 0.804 & 0.812 & 0.818 & 0.815 & 0.972 \\
GPT-3.5 Turbo & 0.596 & 0.686 & 0.691 & 0.704 & 0.706 & 0.802 & 0.831 & 0.827 & 0.904 \\
GPT-4 & 0.786 & 0.827 & 0.832 & 0.840 & 0.849 & 0.876 & 0.925 & 0.932 & 0.971 \\
Llama 2 70b & 0.376 & 0.481 & 0.564 & 0.591 & 0.575 & 0.655 & 0.600 & 0.672 & 0.837 \\
Llama 2 7b & 0.297 & 0.372 & 0.364 & 0.374 & 0.377 & 0.457 & 0.413 & 0.427 & 0.495 \\
Mistral Large & 0.723 & 0.769 & 0.796 & 0.802 & 0.803 & 0.825 & 0.889 & 0.896 & 0.922 \\
\hline
\end{tabular}
\end{table}

\begin{table}[h!]
\caption{Accuracy by agent and exam for GPT-4}
\label{tab:accuracy-by-agent-and-exam-for-gpt-4}
\centering
\fontsize{7pt}{9pt}
\begin{tabular}{lrrrrrrrrrr}
\hline
\textbf{Agent Name} & \textbf{AQUA-RAT} & \textbf{ARC} & \textbf{Hellaswag} & \textbf{LSAT-AR} & \textbf{LSAT-LR} & \textbf{LSAT-RC} & \textbf{LogiQA} & \textbf{MedMCQA} & \textbf{SAT-English} & \textbf{SAT-Math} \\ 
\hline
Baseline & 0.79 & 0.95 & 0.89 & 0.33 & 0.83 & 0.85 & 0.62 & 0.77 & 0.93 & 0.90 \\
Retry & 0.83 & 0.96 & 0.92 & 0.45 & 0.84 & 0.86 & 0.68 & 0.79 & 0.95 & 0.99 \\
Keywords & 0.85 & 0.97 & 0.91 & 0.45 & 0.85 & 0.88 & 0.69 & 0.81 & 0.94 & 0.97 \\
Advice & 0.87 & 0.98 & 0.92 & 0.45 & 0.84 & 0.88 & 0.71 & 0.84 & 0.94 & 0.97 \\
Instructions & 0.86 & 0.98 & 0.93 & 0.48 & 0.86 & 0.88 & 0.71 & 0.88 & 0.95 & 0.96 \\
Explanation & 0.86 & 0.99 & 0.93 & 0.53 & 0.91 & 0.93 & 0.76 & 0.92 & 0.96 & 0.97 \\
Solution & 0.88 & 1.00 & 0.96 & 0.76 & 0.94 & 0.95 & 0.87 & 0.94 & 0.97 & 0.98 \\
Composite & 0.87 & 1.00 & 0.99 & 0.72 & 0.99 & 0.96 & 0.88 & 0.95 & 0.98 & 0.98 \\
Unredacted & 0.91 & 1.00 & 0.99 & 0.92 & 0.99 & 0.98 & 0.95 & 0.99 & 0.99 & 0.99 \\
\hline
\end{tabular}
\end{table}

\newpage
\subsection{Data}

\begin{figure}[h!]
\centering
\small
\ttfamily
\begin{verbatim}
{
  "source": "arc/arc-challenge-test", 
  "source_id": 1, 
  "topic": "Science", 
  "context": "", 
  "question": "An astronomer observes that a planet rotates faster 
               after a meteorite impact. Which is the most likely effect
               of this increase in rotation?", 
  "choices": {
    "A": "Planetary density will decrease.", 
    "B": "Planetary years will become longer.", 
    "C": "Planetary days will become shorter.", 
    "D": "Planetary gravity will become stronger." }, 
  "answer": "C", 
  "solution":""
}
\end{verbatim}
\normalfont
\caption{Sample of an MCQA problem in JSON-L format – with whitespace added for readability.}
\label{fig:mcqa-problem}
\end{figure}

\begin{figure}[h!]
\centering
\small
\ttfamily
\begin{verbatim}
[System Prompt]
You are an expert in {{topic}}.
Your task is to answer the following multiple-choice questions.
Think step-by-step to ensure you have the correct answer.
Then, answer the question using the following format 'Action: Answer("[choice]")'  
The parameter [choice] is the letter or number of the answer you want to select 
  (e.g. "A", "B", "C", or "D")
For example, 'Answer("C")' will select the choice "C" as the best answer.
You MUST select one of the available choices; the answer CANNOT be "None of the Above".
Be concise in your response but include any essential information.

[Example Problem]
Topic: Geography
Question: What is the capital of the state where Johns Hopkins University is located?
Choices:
  A: Baltimore
  B: Annapolis
  C: Des Moines
  D: Las Vegas

[Example Solution]
Thought: 
Johns Hopkins University is located in Baltimore, Maryland.
The capital of Maryland is Annapolis.
Action: Answer("B")  
\end{verbatim}
\normalfont
\caption{Sample of the answer prompt used by the baseline agent to solve MCQA problems.}
\label{fig:answer-prompt}
\end{figure}

\begin{figure}[H]
\centering
\small
\ttfamily
\begin{verbatim}
[System Prompt]
You are an expert in {{topic}}.
You have incorrectly answered the following multiple-choice question.
Your task is to reflect on the problem, your solution, and the correct answer.
You will then use this information help you answer the same question in the future.
First, explain why you answered the question incorrectly.
Second, list the keywords that describe the type of your errors from most general to most specific. 
Third, solve the problem again, step-by-step, based on your knowledge of the correct answer.
Fourth, create a list of detailed instructions to help you correctly solve this problem in the future.
Finally, create a list of general advice to help you solve similar types of problems in the future.  
Be concise in your response; however, capture all of the essential information.  
For guidance, I will provide you with a single generic example problem and reflection (below).

[Example Input]
Topic: Geography and Math
Question: What is the product of the number of letters contained in the name of the city 
  where Iowa State University is located multiplied by the number of letters 
  contained in the name of the state?
Choices:
    A: 16
    B: 20
    C: 24
    D: 32
Thought:
Iowa State University is located in the city of Ames
ISU is located in the state of Iowa.
Action: Answer("D")
---
Correct Answer: A

[Example Output]
Explanation:
I miscalculated the product of the number of letters in the city and state names. 
The gap in my knowledge was not in geography but in basic arithmetic. 
I knew the correct city and state but made a calculation error.
Error Keywords:
 - Calculation error
 - Arithmetic error
 - Multiplication error
Solution:
Iowa State University is located in the city of Ames
Iowa State University is located in the state of Iowa.
The city name "Ames" contains 4 letters.
The state name "Iowa" contains 4 letters.
The product of 4*4 is 16.
Instructions:
1. Identify the city where the university is located.
2. Identify the state where the university is located.
3. Count the number of letters in the name of the city.
4. Count the number of letters in the name of the state.
5. Multiply the number of letters in the city by the number of letters in the state.
6. Work step-by-step through your mathematical calculations.
7. Double-check your calculations to ensure accuracy.
8. Choose the answer that matches your calculated result.
Advice:
 - Always read the question carefully and understand the problem.
 - Always decompose complex problems into multiple simple steps.
 - Always think through each subproblem step-by-step.
 - Never skip any steps; be explicit in each step of your reasoning.
 - Always double-check your calculations and final answer.
 - Remember that the product of two numbers is the result of multiplying them together, 
     not adding them.
\end{verbatim}
\normalfont
\caption{Sample of the self-reflection prompt used to reflect on incorrectly answered MCQA problems.}
\label{fig:reflection-prompt}
\end{figure}

\begin{figure}[H]
\centering
\small
\ttfamily
\begin{verbatim}
[System Prompt (same)]

[Example Problem (same)]

[Example Solution (same)]

[Reflection Prompt]
Reflection:
You previously answered this question incorrectly.
Then you reflected on the problem, your solution, and the correct answer.
Use your self-reflection (below) to help you answer this question.
Any information that you are not allowed to see will be marked [REDACTED].
{{reflection}}
\end{verbatim}
\normalfont
\caption{Sample of the re-answer prompt used by the self-reflecting agents. The system prompt, example problem, and example solution are identical to the answer prompt and thus omitted for clarity.}
\label{fig:re-answer-prompt}
\end{figure}

\end{document}